\documentclass[10pt,twocolumn,letterpaper]{article}

\usepackage{isba}
\usepackage{times}
\usepackage{epsfig}
\usepackage{graphicx}
\usepackage{amsmath}
\usepackage{amssymb}
\usepackage{multirow}

\usepackage{caption} 
\isbafinalcopy 

\ifisbafinal\fi

\begin{document}

\title{Maximum Entropy Binary Encoding for Face Template Protection}

\author{Rohit Kumar Pandey \qquad Yingbo Zhou \qquad Bhargava Urala Kota \qquad Venu Govindaraju \\
University at Buffalo, SUNY\\
{\tt\small $\left\{\text{rpandey, yingbozh, buralako, govind}\right\}$@buffalo.edu}
}

\maketitle

\thispagestyle{empty}

\vspace{-100pt} 
\begin{abstract}

   In this paper we present a framework for secure identification using deep neural networks, and apply it to the task of template protection for face authentication. We use deep convolutional neural networks (CNNs) to learn a mapping from face images to maximum entropy binary (MEB) codes. The mapping is robust enough to tackle the problem of exact matching, yielding the same code for new samples of a user as the code assigned during training. These codes are then hashed using any hash function that follows the random oracle model (like SHA-512) to generate protected face templates (similar to text based password protection). The algorithm makes no unrealistic assumptions and offers high template security, cancelability, and state-of-the-art matching performance. The efficacy of the approach is shown on CMU-PIE, Extended Yale B, and Multi-PIE face databases. We achieve high ($\sim95\%$) genuine accept rates (GAR) at zero false accept rate (FAR) with up to $1024$ bits of template security.
\end{abstract}

\section{Introduction}

Authentication on the basis of ``who we are" instead of ``something we possess" or ``something we remember", offers convenience and often, stronger system security. One of the important factors related to making biometric passwords as widespread as text based ones is that of template protection. Text based password authentication provides strong template protection whereas biometric data generally suffers from lesser protection due to difficulties in exact matching. Given the sensitive nature of biometric data, algorithms that provide the same level of template security without compromising on matching accuracy would be ideal.

A typical password authentication system would use a sample of the user's password to extract and store a template from it. It is desirable that this template is stored in a protected and cancelable manner for the purpose of system security. During authentication, a new template is extracted from the presented password and matched to the stored template. Depending on the matching score, access is granted or denied. In the case of text based passwords, a one way non-invertible transform (i.e.\ a hash) of it is stored as the template. During verification, a password is entered and its hash value is calculated. The hash is compared with the stored hash and if the two strings matched exactly, their hashes would match as well, and access would be granted. In such a scenario, the stored hash reveals no information about the original password (protection) and also, if the password is compromised, it can be changed and a new password can be registered (cancelability).

This kind of security would be ideal for biometric based authentication as well but, unlike text passwords, biometric modalities lack two important aspects. 1) They rarely match exactly between different readings, and 2) they cannot be changed if compromised. Thus, the objective of cancelable biometrics approaches is to extract template from biometric modalities that are 1) protected i.e.\ given the template, it should be infeasible to extract any information about the original modality, and 2) cancelable i.e.\ if compromised, it should be possible to extract a new template from the same modality.

\subsection{Contribution}
We tackle these objectives by using a deep convolutional neural network (CNN) to learn a robust mapping of face classes to maximum entropy binary (MEB) codes. The mapping is robust enough to tackle the problem of exact matching, yielding the same code for new samples of a user as the code assigned during training. This exact matching enables us to store a hash of the code as the template of the user. The hash function used could be any function that follows the random oracle model, and in our case we choose SHA-512 since it is the current standard for string based passwords and offers strong security. Once hashed, the template has no correlation with the code assigned to the user. Furthermore, the codes assigned to users are bit-wise randomly generated and thus, possess maximum entropy, and have no correlation with the original biometric modality (the user's face). These properties make attacks on the template very difficult, leaving brute force attacks as the only feasible option. Cancelability is achieved by changing the codes assigned to users and re-learning the mapping.

Exploiting the large learning capacity of the CNN with powerful regularization, we also achieve state-of-the-art matching performance on PIE, Extended Yale B and Multi-PIE databases. Note that, in this work, we focus on the use-case of using faces as passwords and thus, validate our results on data collected in controlled environments.

\subsection{Related Work}
A variety of template protection algorithms have been applied to faces. Schemes that used cryptosystem based approaches include Fuzzy commitment schemes by Ao and Li~\cite{ao2009near}, Lu \etal~\cite{lu2009face} and Van Der Veen \etal~\cite{van2006face}, and fuzzy vault by  Wu and Qiu~\cite{wu2010transforming}. In general, the fuzzy commitment schemes suffered from limited error correcting capacity or short keys. In Fuzzy vault schemes the data is stored in the open between chaff points, and this also causes an overhead in storage space. Some quantization schemes were used by Sutcu~\etal~\cite{sutcu2007protecting,sutcu2005secure} to generate somewhat stable keys. There were also several works that combine the face data with user specific keys. These include combination with a password by Chen and Chandran~\cite{chen2007biometric}, user specific token binding by Ngo~\etal~\cite{ngo2006biometric,teoh2006random,teoh2004personalised}, biometric salting by Savvides~\etal~\cite{savvides2004cancelable}, and user specific random projection schemes by Teoh and Yuang~\cite{teoh2007cancelable} and Kim and Toh~\cite{kim2007method}. Hybrid approaches that combine transform based cancelability with cryptosystem based security like \cite{feng2010hybrid} have also been proposed but give out user specific information to generate the template creating possibilities of masquerade attacks. Pandey and Govindraju \cite{pandey2015secureface} proposed a security centric scheme that used features extracted from local regions of the face to obtain exact matching and thus, benefited from the security of hash functions. Although more secure, the matching accuracy of the scheme suffered and the feature space being hashed was not uniformly distributed.

On the image recognition side, deep CNNs algorithms like Deepface \cite{taigman2014deepface} have shown exceptional performance and hold the current state-of-the-art results for face recognition. There is also some recent work that seeks to map data to binary codes using deep neural networks like \cite{erin2015deep}. Although mapping to binary codes (or learning hash functions) in this manner may seem similar to our approach, these methods are fundamentally different from what we are trying to achieve. Algorithms such as \cite{erin2015deep} seek to learn a natural binary representation of the data and thus, the binary codes they map to are correlated to the data distribution. Our MEB codes have no correlation to the original data distribution. This gives us the template security we seek, but also makes it a more challenging problem since the mapping function we seek to learn is more complex.

\section{Algorithm}

\begin{figure}
\centering
  \includegraphics[width=\linewidth]{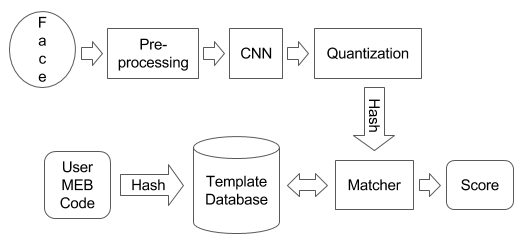}%
  \caption{Overview of the algorithm.}
  \label{fig:overview}
\end{figure}

In this section of the paper we describe the individual components of our architecture in more detail. An overview of the algorithm is shown in Figure \ref{fig:overview}.

\subsection{Convolutional Neural Networks}
Convolutional neural networks (CNNs) \cite{Lecun98gradient-basedlearning} are biologically inspired models, which contain three basic components: convolution, pooling and fully connected layers.  In the convolution layer one tries to learn a filter bank given input feature maps. The input of a convolution layer is a 3D tensor with $d$ number of 2D feature maps of size $n_1\times n_2$.  Let $x_{ijk}$ denote the component at row $j$ and column $k$ in the $i$th feature map, and we use $x_i^{(l)}$ to denote the complete $i$th feature map at layer $l$.  If one wants to learn $h_f$ set of filters of size $f_1 \times f_2$, the output $x^{(l+1)}$ for the next layer will still be a 3D tensor with $h_f$ number of 2D feature maps of size $(n_1-f_1+1) \times (n_2-f_2+1)$.  More formally, the convolution layer computes the following:
\begin{equation}
x_j^{(l+1)} = s(\sum_i F^{(l)}_{ij} * x_i^{(l)} + b^{(l)}_j)
\end{equation}
where $F^{(l)}_{ij}$ denotes the filter that connects feature map $x_i$ to output map $x_j^{(l)}$ at layer $l$, $b^{(l)}_j$ is the bias for the $j$th output feature map, $s(\cdot)$ is some element-wise non-linearity function and $*$ denotes the discrete 2D convolution. 

The pooling (or subsample) layer takes a 3D feature map and tries to down-sample/summarize the content with less spatial resolution. Pooling is commonly done for every feature map independently and with non-overlapping windows.  An intuition of such operation is to have some built in invariance against small translations as well as reduce the spatial resolution and thus save computation for the upper layers.  
For average (mean) pooling, the output will be the average value inside the pooling window, and for max pooling the output will be the maximum value inside the pooling window.  

The fully connected layer connects all the input units from the lower layer $l$ to all the output units in the next layer $l+1$.  In more detail, the next layer output is calculated by:
\begin{equation}
x^{(l+1)} = s(W^{(l)}x^{(l)} + b^{(l)})
\end{equation}
where $x^{(l)}$ is the vectorized input from layer $l$, $W^{(l)}$ and $b^{(l)}$ are the parameters of the fully connected layers at layer $l$.  

A CNN is commonly composed of several stacks of convolution and pooling layers followed by a few fully connected layers. The last layer is normally associated with some loss to provide training signals, and the training for CNN can be done by doing gradient descent on the parameters with respect to the loss. For example, in classification the last layer is normally a softmax layer and cross entropy loss is calculated against the 1 of K representation of the class labels.  In more detail, let $x^{(L)} = Wx^{(L-1)}+b$ be the pre-activation of the last layer, $\mathbf{t}$ denotes the final output and $t_k$ the $k$th component of $\mathbf{t}$, and $\mathbf{y}$ denote the target 1 of K vector and $y_k$ the $k$th dimension of that vector, then
\begin{align}
t_k &= \frac{\exp\{x_k^{(L)}\}}{\sum_j \exp\{x_j^{(L)}\}}\\
L(\mathbf{t}, \mathbf{y}) &= \sum_j y_j \log t_j
\end{align}
where $L$ is the loss function.

\subsection{Maximum Entropy Binary Codes}
Our first step of training is to assign unique codes to each user to be enrolled. From a template security point of view, these codes should ideally possess two properties. First, they should posses high entropy. Since a hash of these codes is the final protected template, the higher the entropy of the codes, the larger the search space for a brute force attack would be. In order to make brute force attacks in the code domain infeasible, we use binary codes with a minimum size $K=256$ bits  and experiment with values up to $K=1024$ bits. The second desirable property of the codes is that they should not be correlated with the original biometric modality. Any correlation between the biometric samples and the secure codes can be exploited by an attacker to reduce the search space during a brute force attack. One example to illustrate this can be to think of binary features extracted from faces. Even though the dimensionality of the feature vector may be high, given the feature extraction algorithm and type of data, the number of possible values the vector can take is severely reduced. In order to prevent such reduction of entropy, the codes we used are bit-wise randomly generated and have no correlation with the original biometric samples. This makes the space to be hashed truly uniformly distributed. More precisely, let $c_i \sim \mathbf{B}(1, 0.5)$ be the binary variable for each bit of the code, where $\mathbf{B}(1, 0.5)$ is the maximum entropy Bernoulli distribution, and the resultant MEB code with $K$ independen bits is thus $\mathbf{C} = [c_1, c_2, \ldots, c_K]$. We denote the code for user $u$ by $\mathbf{C}_u$.

\subsection{Learning the Mapping}
In order to learn a robust mapping of a user's face samples to the codes, we make some modifications to the CNN training procedure. The 1 of K encoding of the class labels is replaced by the MEB codes $\mathbf{C}_u$ assigned to each user. Since we now want several bits of the network output to be one instead of a single bit, we use sigmoid activation instead of softmax. In more detail: 
\begin{align}
t_k &= \frac{1}{1+\exp\{-x_j^{(L)}\}}\\
L(\mathbf{t}, \mathbf{C}) &= \sum_j \{ c_j \log t_j + (1-c_j) \log(1-t_j) \}
\end{align}
where $t_k$ is the $k$th output from the last layer and $L$ is the binary cross entropy loss.

\subsubsection{Data Augmentation}
\label{sec:aug}
Deep learning algorithms generally require a large number of training samples whereas, training samples are generally limited in the case of biometric data. In order to magnify the number of training samples per user, we perform the following data augmentation. For each training sample of size $m \times m$ we extract all possible crops of size $n \times n$. Each crop is also flipped along its vertical axis yielding a total of $ 2 \times (m-n+1)\times(m-n+1)$ crops. The crops are then re-sized back to $m \times m$ and used for training the CNN.

\subsubsection{Regularization}
The large learning capacity of deep neural networks comes with the inherent risk of over-fitting. The number of parameters in the network are often enough to memorize the entire training set, and the performance of such a network does not generalize to new data. In addition to general concerns, mapping to MEB codes is equivalent to learning a highly complex function, where each dimension of the function output can be regarded as an arbitrary binary partition of the classes. This further increases the risk of over-fitting and powerful regularization techniques need be employed to achieve good matching performance.

We apply dropout \cite{hinton2012dropout} on all fully connected layers with 0.5 probability of discarding one hidden activation.  Dropout is a very effective regularizer and can also be regarded as training an ensemble of an exponential number of neural networks that share the same parameters, therefore reducing the variance of the resulting model. 

\subsection{Protected Template}
Even though MEB codes assigned to each user have no correlation with the original samples, another step of taking a hash of the code is required to generate the protected template. Given the parameters of the network, it is not possible to entirely recover the original samples from the code (due to the max pooling operation in the forward pass of the network) but, some information is leaked. Using a hash digest of the code as the final protected template prevents any information leakage. The hash function used can be any function that follows the random oracle model. For our experiments we utilized SHA-512, yielding the final protected template $\mathbf{T}_u = \text{SHA512}(\mathbf{C}_u)$.

During verification, a new sample of the enrolled user is fed through the network to get the network output $\mathbf{y}_{out} = \mathbf{t}$. We then binarize this output via a simple thresholding operation yielding the code for the sample $\mathbf{s}_{out} = [s_1, s_2, \ldots, s_K]$, where $s_i = \mathbf{1}(t_i > 0.5)$ and $\mathbf{1}(\cdot)$ is the indicator function. At this point, the SHA-512 hash of the code, $\mathbf{H}_{out} = \text{SHA512}(\mathbf{s}_{out})$ could be taken and compared with the stored hash $\mathbf{T}_u$ for the user. Due to the exact matching nature of the framework, this would yield a matching score of true/false nature. This is not ideal for a biometric based authentication system since it is desirable to obtain a tunable score in order to adjust the false accept (FAR) and false reject rates (FRR). In order to obtain an adjustable score, several crops and their flipped counterparts are taken for the new sample (in the manner described in Section \ref{sec:aug}) and $\mathbf{H}_{out}$ is calculated for each one, yielding a set of hashes $\mathbb{H}$. We define the final matching score as the number of $\mathbf{H}_{outs}$ in $\mathbb{H}$ that match the stored template, scaled by the cardinality of $\mathbb{H}$. Thus, the score for matching against user $u$ is given by,
\begin{equation}
\label{eq:score}
score = \dfrac{\sum_{\mathbf{H}_i\in \mathbb{H}}\mathbf{1}(\mathbf{H}_i = \mathbf{T}_u)}{\left\vert{\mathbb{H}}\right\vert}
\end{equation}
Now the score can be set to achieve the desired value of FAR/FRR. Note that, the framework provides the flexibility to work in both verification and identification modes. For identification $\mathbb{H}$ can be matched against templates of all the users stored in the database.

\section{Experiments}
We now describe the databases, evaluation protocols, and specifics of the parameters used for experimental evaluation.

\subsection{Databases}
In this study we tackle the the problem of using faces as passwords and thus, choose face databases that have been collected in controlled environments for experimentation. We use evaluation protocols including variations in lighting, session and pose that would be typical to the application. 

The CMU PIE \cite{sim2002cmu} database consists of 41,368 images of 68 people under 13 different poses, 43 different illumination conditions, and with 4 different expressions. We use 5 poses (c27, c05, c29, c09 and c07) and all illumination variations for our experiments. 10 images are randomly chosen for training and the rest are used for testing.

The extended Yale Face Database B \cite{GeBeKr01} contains 2432 images of 38 subjects with frontal pose and under different illumination variations. We use the cropped version of the database for our experiments. Again, we use 10 randomly selected images for training and the rest for testing. 

The CMU Multi-PIE \cite{gross2010multi} face database contains more than 750,000 images of 337 people recorded in 4 different sessions, 15 view points and 19 illumination conditions. We use this database to highlight the algorithm's robustness to changes in session and lighting conditions. We chose two sessions (3 and 4) which had the most number of common users (198) between them. 10 randomly chosen frontal faces from session 3 were used for enrollment and all frontal faces from session 4 were used for verification.

\subsection{Evaluation Metrics}
We use the genuine accept rate (GAR) at 0 false accept rate (FAR) as the evaluation metric. We also report the equal error rate (EER) as an alternative operating point for the system. Since the train-test splits we use are randomly generated, we report the mean and standard deviation of the results for 10 different splits.

\subsection{Experimental Parameters}
We use the same training procedure for all databases. The CNN architecture that we used is as follows: two convolutional layers of $32$ filters of size $7\times 7$ and $64$ filters of size $7\times 7$, each followed by max pooling layers of size $2\times 2$. The convolutional and pooling layers are followed by two fully connected layers of size $2000$ each, and finally the output. We use rectifier activation function $s(x) = \max (x, 0)$ for all layers, and apply dropout with $0.5$ probability of discarding activations to both fully connected layers.

MEB codes of dimensionality $K=256, 1024$ are assigned to each user. All training images are re-sized to $m \times m = 64 \times 64$ and roughly aligned using eye center locations. For augmentation we use $n \times n = 57 \times 57$ crops yielding $64$ crops per image. Each crop is also illumination normalized using the algorithm in \cite{tan2010enhanced}. We train the network by minimizing the cross-entropy loss against user codes for $20$ epochs using mini-batch stochastic gradient descent with a batch size of $200$. 5 of the training samples are initially used for validation to determine the mentioned training parameters. Once the network is trained, the SHA-512 hashes of the codes are stored as the protected templates and the original codes are purged. During verification, crops are extracted from the new sample, pre-processed, and fed through the trained network. Finally, the SHA-512 hash of each crop is calculated and matched to the stored template, yielding the matching score in Equation \ref{eq:score}.

\begin{figure*}[!tbp]
\centering
  \begin{tabular}{ccc}
   \includegraphics[width=0.3\linewidth]{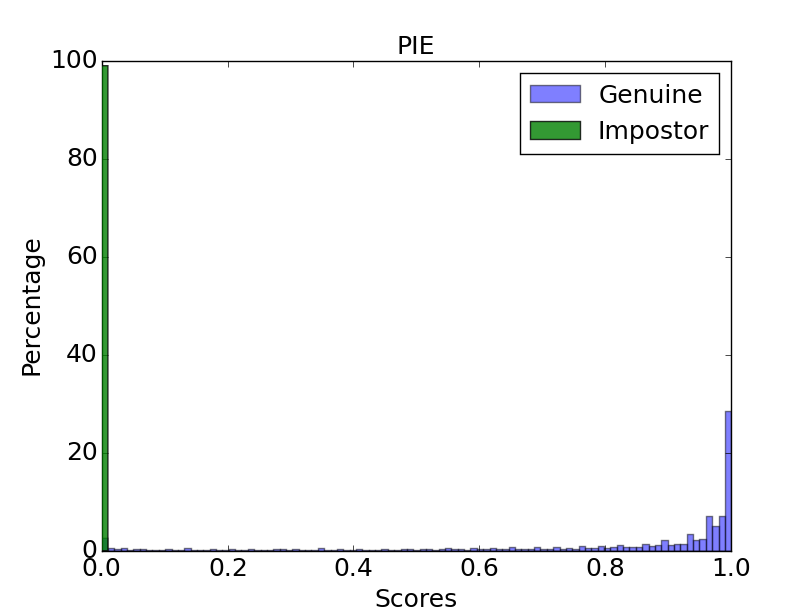}  &
   \includegraphics[width=0.3\linewidth]{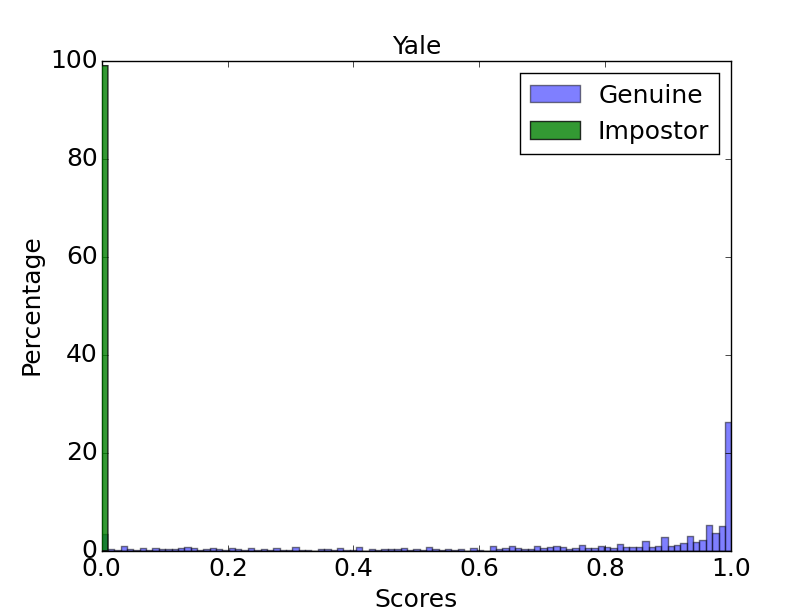} &
   \includegraphics[width=0.3\linewidth]{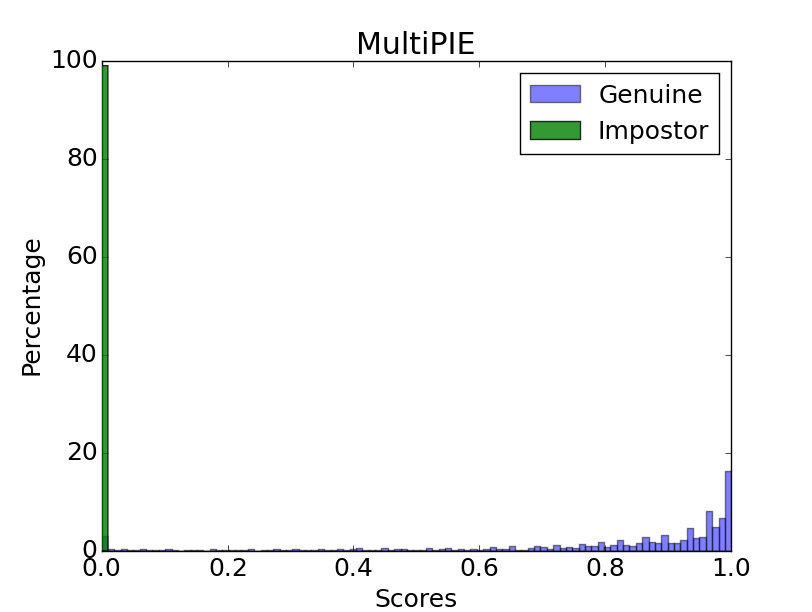}
  \end{tabular}
  \caption{Genuine and imposter distributions from PIE (left), Yale (mid) and Multi-PIE (right) databases.}
\end{figure*}

\subsection{Results}
The results of our experiments are shown in Table \ref{tab:results}. We report the mean and standard deviation of GAR at zero FAR, and EER for the 10 different train-test splits at bits of security (BoS) or $K=256,1024$. We achieve GARs up to $\sim90\%$ on PIE, $\sim96\%$ on Yale, and $\sim97\%$ on Multi-PIE with up to $1024$ bits of security at the strict operating point of zero FAR. During experimentation we observed that our results were stable with respect to $K$, making the parameter selectable purely on the basis of desired template security. A comparison of our results to other face template protection algorithms on the PIE database is shown in Table \ref{tab:comparison}. Our algorithm offers significantly higher template security with true 1024 BoS due to the MEB codes. In terms of matching performance we outperform \cite{feng2010hybrid}, which offers acceptable BoS, and are comparable to \cite{feng2012binary}, which lacks in adequate BoS for protection against brute force attacks.

\begin{table}[!tbp]
\centering
\caption{Verification results obtained from various datasets.}
\label{tab:results}
\begin{tabular}{|c|c|c|c|}
\hline
Database & BoS (K) & GAR@0FAR & EER \\ \hline
\multirow{2}{*}{PIE} & $256$ & $93.22 \pm 2.61\%$  & $1.39 \pm 0.20\%$  \\ \cline{2-4} 
 & $1024$ & $90.13 \pm 4.30\%$  & $1.14 \pm 0.14\%$  \\ \hline
\multirow{2}{*}{Yale} & $256$ & $96.74 \pm 1.35\%$   & $0.93 \pm 0.18\%$  \\ \cline{2-4} 
 & $1024$ & $96.49 \pm 2.30\%$  & $0.71 \pm 0.17\%$  \\ \hline
\multirow{2}{*}{Multi-PIE} & $256$ & $95.93 \pm 0.55\%$  & $1.92 \pm 0.27\%$  \\ \cline{2-4} 
 & $1024$ & $97.12 \pm 0.45\%$  & $0.90 \pm 0.13\%$   \\ \hline
\end{tabular}
\end{table}

\begin{table}[!tbp]
\centering
\caption{Performance comparison with other algorithms on PIE dataset.}
\label{tab:comparison}
\begin{tabular}{|c|c|c|c|}
\hline
Method & BoS (K) & GAR@1FAR & EER \\ \hline
Hybrid Approach \cite{feng2010hybrid} & $210$ & $90.61$\% & $6.81\%$ \\ \hline
BDA \cite{feng2012binary} & $76$ & $96.38$\% & $-$ \\ \hline
MEB Encoding & $\mathbf{1024}$ & $\mathbf{97.59\%}$ & $\mathbf{1.14\%}$ \\ \hline
\end{tabular}
\end{table}

\section{Security Analysis}
We analyze the security of the system in a stolen template scenario. The attacker has possession of the protected templates, knowledge of the template generation algorithm, and the CNN parameters. Given these, the attacker's goal is to extract information about the original biometric of the users. The only assumption we make is that the hash function we use follows the random oracle model. Due to this, given the hash digests, the attacker cannot extract any information about the MEB codes assigned to the users. This removes the possibility of using the CNN parameters to reverse engineer the face from the secure codes. Now, the only way in which the attacker can get the codes is by brute forcing through all possible values the codes can take, hash each one, and compare to the hashed templates. Since the minimum code length we use is $256$ bits, the search space is of the order of $2^{256}$ or larger, making brute force attacks computationally infeasible.

\begin{figure}
\centering
  \includegraphics[width=0.8\linewidth]{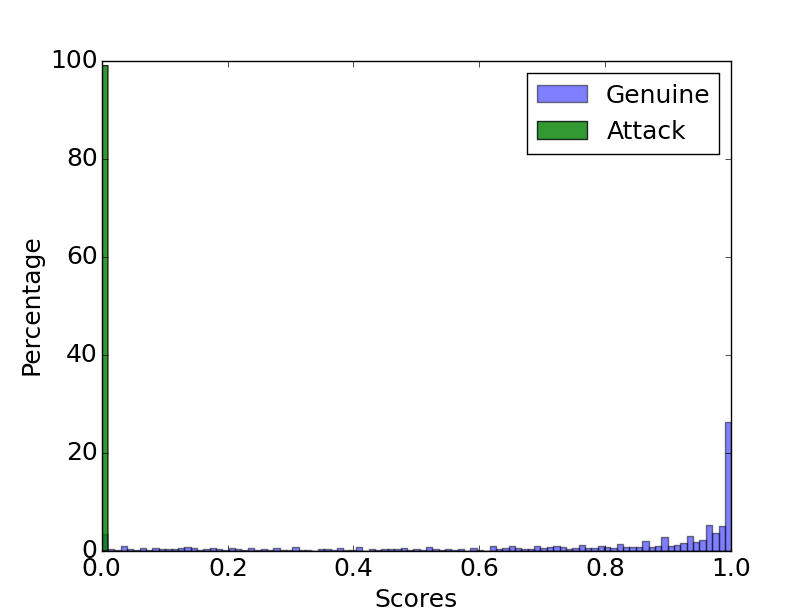}%
  \caption{Genuine and imposter distributions for attacks in the input space.}
  \label{fig:unseen}
\end{figure}

Another possibility of an attack could be brute force in the input domain i.e.\ feed random noise or faces into the network and hope for a match. This comes down to the question of the entropy of faces in general, which is beyond the scope of this paper but, we do empirically analyze the behavior of the network under such attacks. So far, the imposter scores have been calculated using other enrolled users. We now analyze the score distribution when face samples that have not been seen by the network are fed to it. For this experiment, we enroll samples from Extended Yale B Database and use all the faces in Multi-PIE as imposter samples. In addition to unseen faces, we also feed 1 million samples of random noise through the network. The results of this experiment are shown in Figure \ref{fig:unseen} with the attack distribution representing the scores for faces from Multi-PIE and the random noise. It can be seen that the scores for the attack data are always zero and well separated from the genuine scores, empirically verifying the security of the system to attacks in the input space.

\section{Conclusion and Future Work}
We presented a template protection algorithm which achieves provable security by using MEB codes to address the issue of uniformity, and relying on the strength of standard hash functions. We achieved high ($\sim95\%$) GARs at the strict operating point of zero FAR and showed that the exceptional performance of deep CNNs can be utilized to minimize loss of matching accuracy in template protection algorithms. The current work deals with the problem of using faces as passwords in controlled environments, and we plan to extend our results to faces in uncontrolled environments, other biometric modalities, and broader applications like Microsoft Windows picture passwords. Our future efforts also seek to make a formal analysis of our algorithm from an information theoretic perspective.




{\small
\bibliographystyle{ieee}
\bibliography{egbib}

\begin{thebibliography}{10}\itemsep=-1pt

\bibitem{ao2009near}
M.~Ao and S.~Z. Li.
\newblock Near infrared face based biometric key binding.
\newblock In {\em Advances in Biometrics}, pages 376--385. Springer, 2009.

\bibitem{chen2007biometric}
B.~Chen and V.~Chandran.
\newblock Biometric based cryptographic key generation from faces.
\newblock In {\em Digital Image Computing Techniques and Applications, 9th
  Biennial Conference of the Australian Pattern Recognition Society on}, pages
  394--401. IEEE, 2007.

\bibitem{erin2015deep}
V.~Erin~Liong, J.~Lu, G.~Wang, P.~Moulin, and J.~Zhou.
\newblock Deep hashing for compact binary codes learning.
\newblock In {\em Proceedings of the IEEE Conference on Computer Vision and
  Pattern Recognition}, pages 2475--2483, 2015.

\bibitem{feng2012binary}
Y.~C. Feng and P.~C. Yuen.
\newblock Binary discriminant analysis for generating binary face template.
\newblock {\em Information Forensics and Security, IEEE Transactions on},
  7(2):613--624, 2012.

\bibitem{feng2010hybrid}
Y.~C. Feng, P.~C. Yuen, and A.~K. Jain.
\newblock A hybrid approach for generating secure and discriminating face
  template.
\newblock {\em Information Forensics and Security, IEEE Transactions on},
  5(1):103--117, 2010.

\bibitem{GeBeKr01}
A.~Georghiades, P.~Belhumeur, and D.~Kriegman.
\newblock From few to many: Illumination cone models for face recognition under
  variable lighting and pose.
\newblock {\em IEEE Trans. Pattern Anal. Mach. Intelligence}, 23(6):643--660,
  2001.

\bibitem{gross2010multi}
R.~Gross, I.~Matthews, J.~Cohn, T.~Kanade, and S.~Baker.
\newblock Multi-pie.
\newblock {\em Image and Vision Computing}, 28(5):807--813, 2010.

\bibitem{hinton2012dropout}
G.~E. Hinton, N.~Srivastava, A.~Krizhevsky, I.~Sutskever, and R.~R.
  Salakhutdinov.
\newblock Improving neural networks by preventing co-adaptation of feature
  detectors.
\newblock {\em arXiv preprint arXiv:1207.0580}, 2012.

\bibitem{kim2007method}
Y.~Kim and K.-A. Toh.
\newblock A method to enhance face biometric security.
\newblock In {\em Biometrics: Theory, Applications, and Systems, 2007. BTAS
  2007. First IEEE International Conference on}, pages 1--6. IEEE, 2007.

\bibitem{Lecun98gradient-basedlearning}
Y.~Lecun, L.~Bottou, Y.~Bengio, and P.~Haffner.
\newblock Gradient-based learning applied to document recognition.
\newblock In {\em Proceedings of the IEEE}, pages 2278--2324, 1998.

\bibitem{lu2009face}
H.~Lu, K.~Martin, F.~Bui, K.~Plataniotis, and D.~Hatzinakos.
\newblock Face recognition with biometric encryption for privacy-enhancing
  self-exclusion.
\newblock In {\em Digital Signal Processing, 2009 16th International Conference
  on}, pages 1--8. IEEE, 2009.

\bibitem{ngo2006biometric}
D.~C. Ngo, A.~B. Teoh, and A.~Goh.
\newblock Biometric hash: high-confidence face recognition.
\newblock {\em Circuits and Systems for Video Technology, IEEE Transactions
  on}, 16(6):771--775, 2006.

\bibitem{pandey2015secureface}
R.~K. Pandey and V.~Govindaraju.
\newblock Secure face template generation via local region hashing.
\newblock In {\em Biometrics (ICB), 2015 International Conference on}, pages
  1--6. IEEE, 2015.

\bibitem{savvides2004cancelable}
M.~Savvides, B.~V. Kumar, and P.~K. Khosla.
\newblock Cancelable biometric filters for face recognition.
\newblock In {\em ICPR 2004}, volume~3, pages 922--925. IEEE, 2004.

\bibitem{sim2002cmu}
T.~Sim, S.~Baker, and M.~Bsat.
\newblock The cmu pose, illumination, and expression (pie) database.
\newblock In {\em Automatic Face and Gesture Recognition, 2002. Proceedings.
  Fifth IEEE International Conference on}, pages 46--51. IEEE, 2002.

\bibitem{sutcu2007protecting}
Y.~Sutcu, Q.~Li, and N.~Memon.
\newblock Protecting biometric templates with sketch: Theory and practice.
\newblock {\em Information Forensics and Security, IEEE Transactions on},
  2(3):503--512, 2007.

\bibitem{sutcu2005secure}
Y.~Sutcu, H.~T. Sencar, and N.~Memon.
\newblock A secure biometric authentication scheme based on robust hashing.
\newblock In {\em Proceedings of the 7th workshop on Multimedia and security},
  pages 111--116. ACM, 2005.

\bibitem{taigman2014deepface}
Y.~Taigman, M.~Yang, M.~Ranzato, and L.~Wolf.
\newblock Deepface: Closing the gap to human-level performance in face
  verification.
\newblock In {\em Computer Vision and Pattern Recognition (CVPR), 2014 IEEE
  Conference on}, pages 1701--1708. IEEE, 2014.

\bibitem{tan2010enhanced}
X.~Tan and B.~Triggs.
\newblock Enhanced local texture feature sets for face recognition under
  difficult lighting conditions.
\newblock {\em Image Processing, IEEE Transactions on}, 19(6):1635--1650, 2010.

\bibitem{teoh2007cancelable}
A.~Teoh and C.~T. Yuang.
\newblock Cancelable biometrics realization with multispace random projections.
\newblock {\em Systems, Man, and Cybernetics, Part B: Cybernetics, IEEE
  Transactions on}, 37(5):1096--1106, 2007.

\bibitem{teoh2006random}
A.~B. Teoh, A.~Goh, and D.~C. Ngo.
\newblock Random multispace quantization as an analytic mechanism for
  biohashing of biometric and random identity inputs.
\newblock {\em Pattern Analysis and Machine Intelligence, IEEE Transactions
  on}, 28(12):1892--1901, 2006.

\bibitem{teoh2004personalised}
A.~B. Teoh, D.~C. Ngo, and A.~Goh.
\newblock Personalised cryptographic key generation based on facehashing.
\newblock {\em Computers \& Security}, 23(7):606--614, 2004.

\bibitem{van2006face}
M.~Van Der~Veen, T.~Kevenaar, G.-J. Schrijen, T.~H. Akkermans, F.~Zuo, et~al.
\newblock Face biometrics with renewable templates.
\newblock In {\em Proceedings of SPIE}, volume 6072, page 60720J, 2006.

\bibitem{wu2010transforming}
Y.~Wu and B.~Qiu.
\newblock Transforming a pattern identifier into biometric key generators.
\newblock In {\em ICME 2010}, pages 78--82. IEEE, 2010.

\end{thebibliography}
}

\end{document}